\documentclass[letterpaper]{article}
\usepackage[preprint]{aaai2027}
\usepackage[hyphens]{url}
\usepackage{graphicx}
\usepackage{natbib}
\usepackage{caption}
\usepackage{booktabs}
\usepackage{amsmath,amssymb}
\usepackage{array}
\usepackage{pdfpages}
\newcolumntype{L}[1]{>{\raggedright\arraybackslash}m{#1}}
\newcolumntype{C}[1]{>{\centering\arraybackslash}m{#1}}
\newlength{\methodcolwidth}
\newlength{\transcriptcolwidth}
\newlength{\metriccolwidth}
\newlength{\transcriptmetriccolwidth}
\newlength{\ablationvariantcolwidth}
\newlength{\ablationmetriccolwidth}
\title{\resizebox{0.98\textwidth}{!}{LLM-Anchored Paralinguistic Enrichment for Alzheimer's Disease Detection}}
\author{
    Xiao Wei\textsuperscript{\rm 1,2},
    Yuqin Lin\textsuperscript{\rm 3},
    Yaru Cao\textsuperscript{\rm 2},
    Jinyu Li\textsuperscript{\rm 1},
    Bin Wen\textsuperscript{\rm 1},\\
    Kai Li\textsuperscript{\rm 2},
    Yueying Chen\textsuperscript{\rm 1},
    Longbiao Wang\textsuperscript{\rm 1,4},
    Jianwu Dang\textsuperscript{\rm 2}
}
\affiliations{
    \textsuperscript{\rm 1}Tianjin Key Laboratory of Cognitive Computing and Application,\\
    College of Intelligence and Computing, Tianjin University, Tianjin, China\\
    \textsuperscript{\rm 2}Shenzhen Institutes of Advanced Technology, Chinese Academy of Sciences, Shenzhen, China\\
    \textsuperscript{\rm 3}College of Computer and Data Science, Fuzhou University, Fuzhou, China\\
    \textsuperscript{\rm 4}Huiyan Technology (Tianjin) Co., Ltd, Tianjin, China\\
    weixiao@tju.edu.cn
}

\begin{document}

\maketitle

\begin{abstract}

Speech-based automatic detection of Alzheimer's disease (AD) provides a non-invasive and scalable approach to early cognitive screening. AD affects both lexical-semantic organization and speech production, including atypical pauses and word elongations. However, existing methods have yet to fully integrate these paralinguistic cues with linguistic content. We propose \textbf{L}LM-\textbf{A}nchored \textbf{P}aralinguistic \textbf{E}nrichment (LAPE), which enriches LLM-derived linguistic representations with paralinguistic cues through three coordinated innovations. The first is prosodic event textualization, which enables the LLM to model pauses and elongations jointly with lexical content by encoding them as explicit markers with bounded duration-aware repetition. The second is lexico-prosodic unitization and chunking, which preserves event identity and magnitude in both modalities by pooling only consecutive word units. The third is text-anchored paralinguistic fusion, which integrates local and utterance-level speech features by using NormGate to normalize and dynamically scale them relative to text. We evaluate LAPE on ADReSS and ADReSSo using participant-level cross-validation and leave-one-subject-out evaluation. LAPE achieves state-of-the-art performance across all four primary settings. Code will be released upon acceptance.
\end{abstract}

\section{Introduction}

Alzheimer's disease (AD) is a progressive neurodegenerative disorder and a leading cause of dementia in older adults. Earlier identification can support timely clinical assessment, disease management, and care planning, motivating the search for accessible, non-invasive, and low-cost screening methods \citep{delafuentegarcia2020review,yang2022review}. Among candidate signals, spontaneous speech is particularly promising because it combines practical accessibility with discriminative value for cognitive status. It can be collected remotely or incorporated into routine cognitive assessment, while capturing both what a person says and how the speech is produced. The former reflects linguistic abilities such as lexical retrieval and semantic organization, while the latter reflects speech-production abilities such as fluency, timing, and articulation; together, they capture complementary aspects of cognitive decline \citep{voleti2020review}. ADReSS \citep{luz2020adress} and ADReSSo \citep{luz2021adresso} provide standard benchmarks for studying this complementarity under two representative input conditions: the former includes official manual transcripts, while the latter provides speech only.

AD detection therefore needs both linguistic and paralinguistic information. For simplicity, we use \emph{paralinguistic cues} to denote speech-production information beyond linguistic content, including prosodic timing, local acoustic realization, and utterance-level voice characteristics. Recent multimodal systems often report stronger text-only than speech-only performance \citep{ying2023multimodal,bang2024llm,cognialign2025,wei2026falad}, suggesting unequal discriminative contributions. The central challenge is therefore to cover heterogeneous paralinguistic cues while accounting for this asymmetry.

\begin{figure}[t]
  \centering
  \includegraphics[width=\columnwidth]{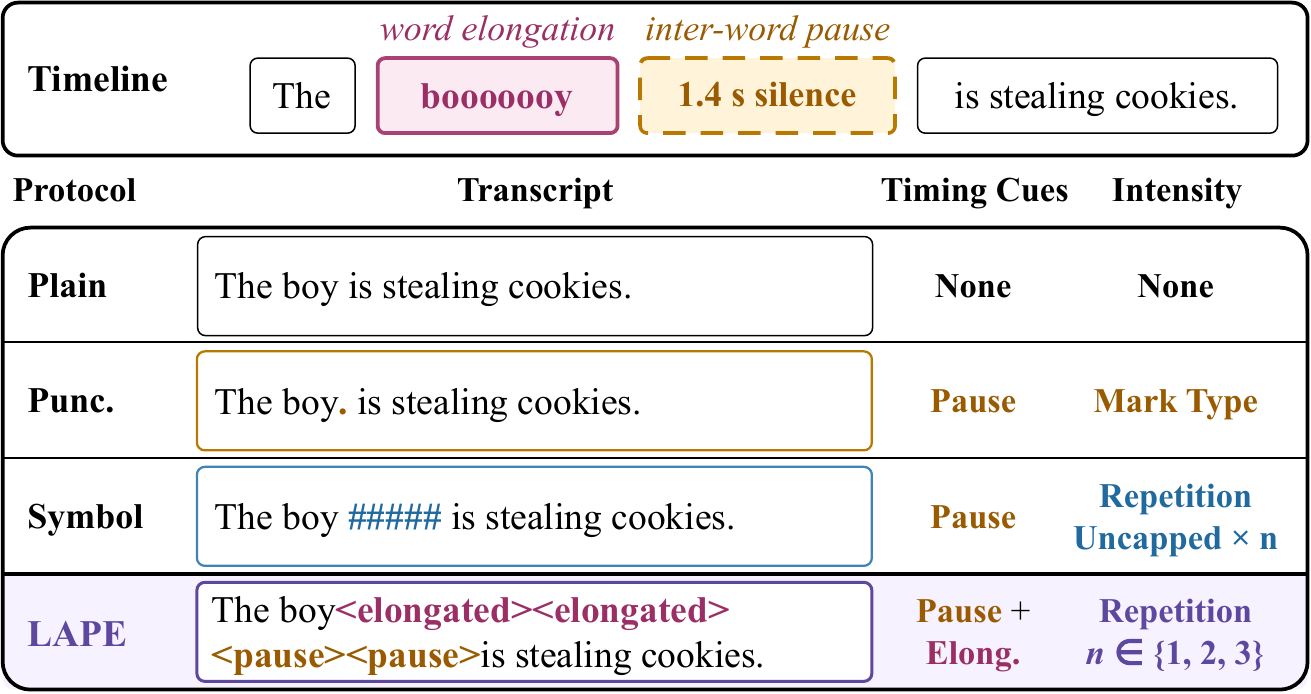}
  \caption{Transcription protocols for the same speech timeline. LAPE places bounded duration-aware repetitions of pause markers before the following word and elongation markers after the host word.}
  \label{fig:transcript-representations}
\end{figure}

Prior work has explored relevant aspects of this challenge along two separate lines. At the transcription level, plain transcripts leave pauses and elongations implicit \citep{zolnoori2023adscreen,voxcog2026,jmir2026markers}. Pause-enhanced protocols map timestamp-derived pauses to punctuation \citep{pauseencoding2024}, but punctuation also serves syntactic and discourse functions, potentially conflating timing with linguistic structure. Figure~\ref{fig:transcript-representations} shows the remaining coverage gap: plain text omits timing events, whereas punctuation-based and uncapped-repetition protocols encode only pauses \citep{yuan2021pauses,lpmaug2026}. At the feature level, multimodal systems combine acoustic and linguistic representations through concatenation, attention, or gating \citep{llmasvc2025,cognialign2025,unsupalignment2026}. Although effective, these systems emphasize fusion operators, leaving local prosodic structure and utterance-level speech features insufficiently coordinated. Together, limited transcript cue coverage and insufficient coordination between local and utterance-level speech features, compounded by unequal modality contributions, motivate a unified design that connects transcription protocol design with feature fusion.

To address these limitations, we propose \textbf{L}LM-\textbf{A}nchored \textbf{P}aralinguistic \textbf{E}nrichment (LAPE) for speech-based AD detection. Rather than treating text and speech symmetrically, LAPE uses LLM-derived linguistic representations as the organizing anchor, with paralinguistic cues serving as structured complements across transcript and speech representations. Our contributions are:

\begin{enumerate}
\def\labelenumi{\arabic{enumi}.}
\item
  \textbf{Prosodic event textualization.} We enrich transcripts with explicit pause and elongation cues using bounded duration-aware marker repetition, enabling the LLM to model them alongside lexical content. To our knowledge, LAPE is the first to make elongations explicit in text.
\item
  \textbf{Lexico-prosodic unitization and chunking.} We organize text and speech into units and introduce chunk aggregation that preserves event identity and magnitude in both modalities by pooling only consecutive word units.
\item
  \textbf{Text-anchored paralinguistic fusion.} We integrate local and utterance-level speech features through NormGate, which normalizes and dynamically scales the speech-derived blocks relative to text.
\item
  \textbf{Extensive evaluation.} Across ADReSS and ADReSSo under CV and LOSO, LAPE achieves state-of-the-art performance in all four settings.
\end{enumerate}

\section{Related Work}

\subsection{Prosody-Enriched Transcript Modeling}

Transcript-based AD detection has progressed from handcrafted indicators to pretrained language models. Plain transcripts primarily preserve lexical content, whereas timestamps expose prosodic timing such as inter-word pauses \citep{luz2020adress,llmprompt2025} and word-level elongations. Earlier studies summarize pause behavior over an entire recording using aggregate statistics or duration distributions \citep{pastoriza2022pausedistribution,ke2024biomarkers}, providing interpretable biomarkers. By contrast, explicit transcript enrichment serializes prosodic events alongside words to preserve event order and lexical context; existing work has focused primarily on pauses.

Explicit pause encodings involve two related design choices. The first concerns how markers enter the language model. Punctuation-based schemes map forced-aligned pauses to commas, periods, or ellipses. \citet{yuan2021pauses} use three pause-duration categories and find that using six bins provides no further improvement, while \citet{pauseencoding2024} adopt a similar three-level scheme in a diagnostic template. Although these schemes exploit pretrained punctuation representations, they may conflate speech timing with syntactic or discourse boundaries. Alternatives use dedicated literal markers \citep{pausefiller2024,lpmaug2026} or added tokens \citep{pausecontext2024}. Literal markers avoid vocabulary modification but may be fragmented by the tokenizer or inherit unrelated semantics; added tokens ensure atomic units but require learning new embeddings from limited downstream data. The second concerns how pause duration is encoded. One approach assigns a different marker to each predefined duration interval, whereas another repeats a single marker at fixed increments, as in LPM-Aug. The former bounds the added length but relies on predefined thresholds and a larger marker inventory; the latter uses one marker type but adds more tokens for longer pauses.

Overall, explicit schemes still focus mainly on inter-word pauses, whereas word-level elongation is usually treated as a continuous duration or acoustic feature \cite{zolnoori2023adscreen,jmir2026markers}. LAPE addresses this gap by jointly textualizing pause events and word-level elongation cues with dedicated event markers instead of standard marks.

\subsection{Acoustic--Linguistic Organization and Fusion}

Acoustic--linguistic fusion in AD detection involves two related considerations: how speech and text are organized and how their relative contributions are handled. Utterance-level methods independently pool both modalities over an entire recording and combine them through concatenation, feature selection, attention, or higher-order interactions \citep{ying2023multimodal,bang2024llm,llmasvc2025,wang2021modular,chatzianastasis2023fusion}. These designs offer broad coverage and straightforward integration but collapse local timing structure, obscuring acoustic variation associated with specific words or prosodic events.

To retain local timing, finer-grained methods establish speech--text relations before fusion. Unsupervised Alignment Feature Fusion learns soft affinities \citep{unsupalignment2026}, while MCI-OTFusion uses an optimal-transport constraint and bidirectional cross-attention \citep{mciotfusion2026}. CogniAlign and FAL-AD instead use explicit temporal anchors with cross-attention \citep{cognialign2025,wei2026falad}. The former two infer correspondences from feature sequences, whereas the latter two rely on timestamps. These methods strengthen local interaction but generally operate on a single speech--text sequence pair, leaving utterance-level acoustic features less explicitly coordinated.

Beyond organization, reported results often show that text-only models outperform their speech-only counterparts, indicating unequal modality contributions. This asymmetry becomes more relevant as language backbones grow stronger. Overall, existing fusion methods do not jointly preserve local prosodic structure, incorporate utterance-level speech features, and account for unequal modality contributions. LAPE uses LLM-derived text as the anchor for coordinating speech evidence across temporal scales.

\section{Methodology: LAPE}

\subsection{Overview}

\begin{figure}[t]
  \centering
  \includegraphics[width=\columnwidth]{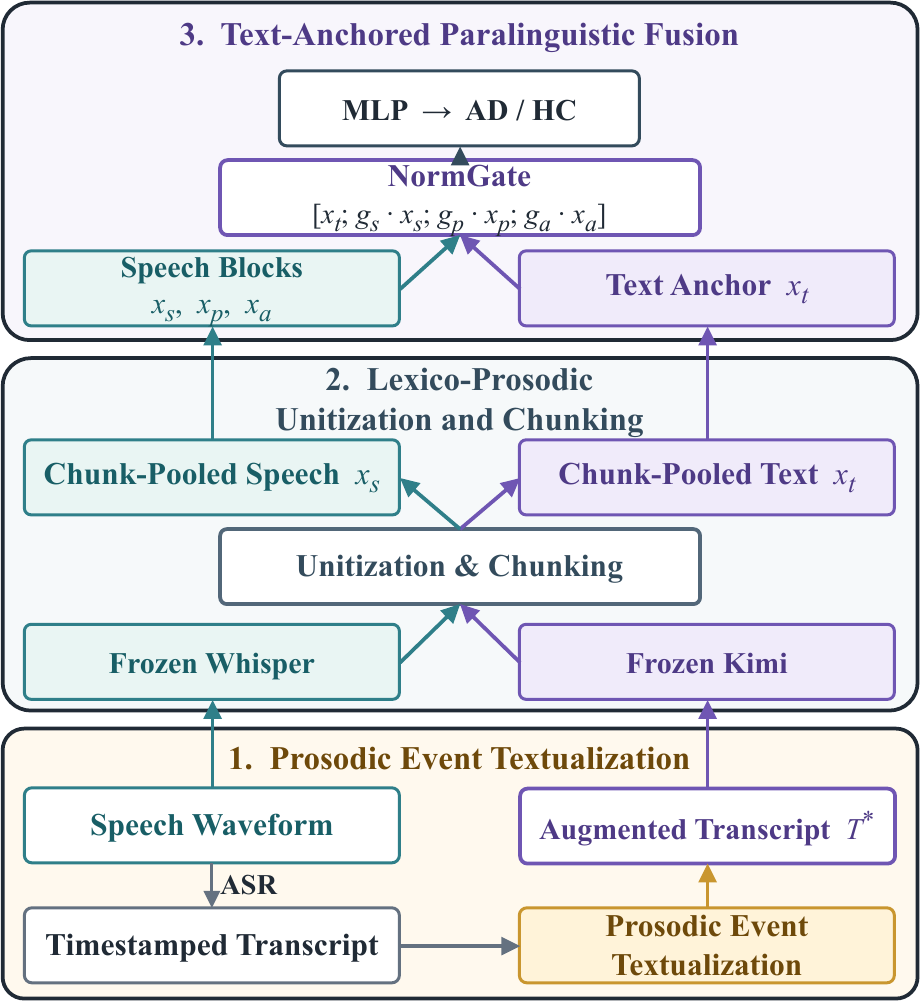}
  \caption{LAPE workflow. Prosodic events are textualized in the transcript; frozen Kimi and Whisper features undergo unitization and event-preserving chunking; NormGate scales speech features relative to a text anchor for classification.}
  \label{fig:lape-workflow}
\end{figure}

Figure~\ref{fig:lape-workflow} presents LAPE for AD detection. Linguistic content often provides the stronger diagnostic signal, while paralinguistic evidence is heterogeneous across temporal scales and should serve as a structured complement. LAPE therefore follows two principles: organizing linguistic and local speech evidence through unitization and event-preserving chunking, and using LLM-derived text as the anchor for integrating local and utterance-level speech evidence.

\subsection{Prosodic Event Textualization}

The first stage augments the transcript. Word timestamps provide two AD-relevant timing cues: inter-word pauses and word duration. Prosodic event textualization writes both as explicit markers in the word sequence, allowing the LLM to model timing with lexical content. Dedicated markers separate speech timing from syntactic and discourse punctuation.

Let the original transcript be the sequence of timestamped words with retained punctuation \(T=((w_i,\pi_i,b_i,e_i))_{i=1}^{L}\), where \(L\) is the number of original words, \(w_i\) is the \(i\)-th word, \(\pi_i\) is the possibly empty punctuation sequence immediately following it, and \(b_i\) and \(e_i\) denote its begin and end times in ms. All timestamp and duration variables in this section use ms. Let \(\Delta_i\) denote the non-negative inter-word pause immediately preceding \(w_i\); we set \(\Delta_1=0\) so that the first word receives an empty pause prefix:

\begin{equation}
\Delta_1=0,
\quad
\Delta_i=\max(0,b_i-e_{i-1}),
\quad i=2,\ldots,L.
\end{equation}
We encode pauses using the marker \texttt{\textless{}pause\textgreater{}}. Following the final protocol, pause durations are mapped to three bins:

\begin{equation}
\phi_p(\Delta_i) = \begin{cases}
\varnothing, & \Delta_i \le 500, \\
\langle\mathtt{pause}\rangle\,\times 1, & 500 < \Delta_i \le 1000, \\
\langle\mathtt{pause}\rangle\,\times 2, & 1000 < \Delta_i \le 2000, \\
\langle\mathtt{pause}\rangle\,\times 3, & \Delta_i > 2000
\end{cases}.
\end{equation}
Because expected word duration depends on syllable count, we compare each word against a fixed syllable-conditioned reference and treat any excess duration as elongation. Let \(n_i=g(w_i)\) denote the estimated syllable count and \(d_i=\max(0,e_i-b_i)\) the observed word duration in ms. The reference duration is:

\begin{equation}
D_{\mathrm{ref}}(n)= \begin{cases} 260, & n=1,\\ 360, & n=2,\\ 450, & n=3,\\ 580, & n=4,\\ 700+150(n-5), & n\geq5 \end{cases}.
\end{equation}
Reference anchors were initialized from syllable-conditioned duration statistics of ADReSSo ASR word alignments, then rounded and smoothed. Sparse data for words with at least five syllables required linear extrapolation for longer words. This fixed table provides a task-specific baseline for coarse duration normalization. The residual \(r_i\) and its \texttt{\textless{}elongated\textgreater{}} encoding are:

\begin{equation}
r_i = \max\left(0,d_i-D_{\mathrm{ref}}(n_i)\right),
\label{eq:elongation-residual}
\end{equation}
\begin{equation}
\phi_e(r_i) = \begin{cases}
\varnothing, & r_i \le 200, \\
\langle\mathtt{elongated}\rangle\,\times 1, & 200 < r_i \le 500, \\
\langle\mathtt{elongated}\rangle\,\times 2, & 500 < r_i \le 800, \\
\langle\mathtt{elongated}\rangle\,\times 3, & r_i > 800
\end{cases}.
\label{eq:elongation-encoding}
\end{equation}

Writing the binned pause and elongation markers into \(T\) while retaining punctuation yields LAPE's augmented transcript, organized as units centered on the original words:

\begin{equation}
\mathcal{A}_i=\phi_p(\Delta_i)w_i\phi_e(r_i)\pi_i,
\quad
T^\star=\mathcal{A}=\bigl(\mathcal{A}_i\bigr)_{i=1}^{L}.
\end{equation}
Index \(i\) links each \(\mathcal{A}_i\) to timestamps \((b_i,e_i)\).

\begin{figure*}[t]
  \centering
  \includegraphics[width=\textwidth,trim=72 0 201 0,clip]{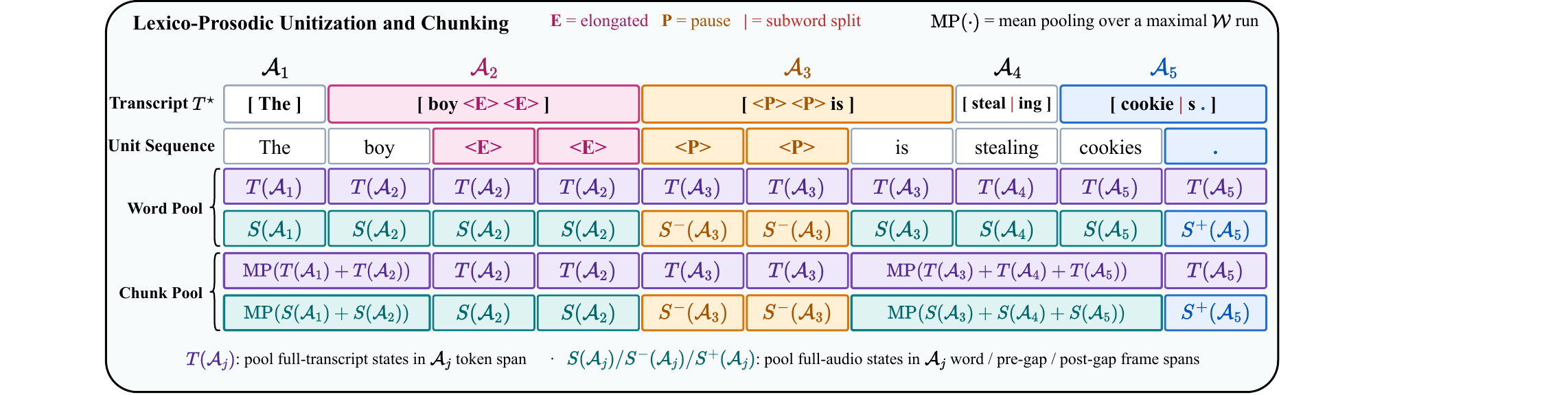}
  \caption{Lexico-prosodic unitization and chunking. Augmented word groups are first unitized into text and speech units. Consecutive word units are then mean-pooled into lexical chunks, while each pause, elongation, and boundary unit remains a singleton chunk; repeated markers therefore preserve event magnitude.}
  \label{fig:unitization-chunking}
\end{figure*}

LAPE feeds \(T^\star\) to Kimi-K2.7-Code\footnote{\url{https://huggingface.co/moonshotai/Kimi-K2.7-Code}.} together with a fixed task instruction; we denote this input by \(\operatorname{Prompt}(T^\star)\). The instruction defines a binary classification task that distinguishes participants with AD from healthy controls (HCs) using cognitive-linguistic information, and explains the duration bins represented by repeated pause and elongation markers. The exact prompt template is given in the supplementary material.

\subsection{Lexico-Prosodic Unitization and Chunking}

Figure~\ref{fig:unitization-chunking} details the unitization and chunking procedure. We use \(L\), \(M\), and \(N\) throughout this subsection for the lengths of the augmented-word-group sequence \(\mathcal{A}\), unit sequence \(U\), and chunk sequence \(\mathcal{C}\), respectively. Prosodic event textualization makes timing cues legible to the LLM, but Kimi still emits subword states and Whisper-small emits frame-level states. Pooling either output directly would weight text by tokenizer fragmentation and blur the word and inter-word intervals that carry timing evidence. We therefore use word timestamps as a deterministic scaffold for a word-level backbone: lexical units organize both modalities, while prosodic events determine which local acoustic intervals are selected and, through repetition, how strongly they contribute. When the marker mappings and punctuation field are empty, \(\mathcal{A}_i\) reduces to \(w_i\); hence \(T^\star\) always contains exactly \(L\) augmented word groups, one for each original word.

Each \(\mathcal{A}_i\) is expanded in transcript order into one word unit and zero or more marker units, providing a basis for text and speech organization and pooling. The word unit corresponds to \(w_i\); every marker instance in \(\phi_p(\Delta_i)\) and \(\phi_e(r_i)\) contributes one pause or elongation unit, respectively; and each punctuation mark in a nonempty \(\pi_i\) contributes a corresponding punctuation-derived boundary unit. Let \(\mathcal{W}\), \(\mathcal{P}\), \(\mathcal{E}\), and \(\mathcal{B}\) denote these four unit classes. Empty fields contribute no additional units, whereas repeated markers remain separate units and preserve event magnitude explicitly. Ordering the expanded units from \(\mathcal{A}_1\) through \(\mathcal{A}_L\) yields \(U=(U_1,\ldots,U_M)\). Each \(U_j\) derives from exactly one \(\mathcal{A}_i\), denoted by \(U_j\in\mathcal{A}_i\).
The frozen Kimi language model processes \(\operatorname{Prompt}(T^\star)\), jointly contextualizing the fixed task instruction and the augmented transcript; only token states \(h_\ell^{(t)}\) corresponding to \(T^\star\) are retained for subsequent aggregation. In parallel, the frozen Whisper encoder \citep{radford2023whisper} processes the speech waveform. Let \(h_\ell^{(s)}\) denote its frame state at timestamp \(\tau_\ell\).

The two modalities follow the same unit order while preserving their distinct evidence structures. A marker derives its textual meaning from the word to which it is attached and the surrounding syntactic context, so all members of \(\mathcal{A}_i\) share a jointly contextualized text span. Its speech realization, however, is localized in time, so speech spans are selected separately for individual units. Specifically, let \(\mathcal{T}_i\) denote the retained Kimi token-position span that covers the original word and all attached markers in \(\mathcal{A}_i\). The selected token states are mean-pooled once, and the resulting vector is assigned to every \(U_j\in\mathcal{A}_i\).

Speech evidence is selected from the acoustic interval indicated by each unit. Let \(\mathcal{S}_j\) denote the Whisper frame span of \(U_j\). We call \([b_i,e_i)\) the \emph{word span} of \(w_i\), which covers the time-frame interval occupied by the current word; \([e_{i-1},b_i)\) and \([e_i,b_{i+1})\) are its \emph{pre-gap} and \emph{post-gap}, namely the intervals separating it from the preceding and following words. For a terminal boundary, we set \(b_{L+1}=e_L\). All intervals are left-closed and right-open; when their endpoints coincide, the nearest valid encoder frame is retained. For units \(U_j\in\mathcal{A}_i\), the word and elongation units use the word span, whereas pause and boundary units use the pre-gap and post-gap, respectively:

\begin{equation}
\mathcal{S}_j=
\begin{cases}
\{\ell\mid\tau_\ell\in[b_i,e_i)\},
& U_j\in\mathcal{W}\cup\mathcal{E},\\
\{\ell\mid\tau_\ell\in[e_{i-1},b_i)\},
& U_j\in\mathcal{P},\\
\{\ell\mid\tau_\ell\in[e_i,b_{i+1})\},
& U_j\in\mathcal{B}
\end{cases}.
\end{equation}
The text and speech representations of each unit are therefore

\begin{equation}
u_j^{(t)}=\frac{1}{|\mathcal{T}_i|}\sum_{\ell\in\mathcal{T}_i}h_\ell^{(t)},
\;
u_j^{(s)}=\frac{1}{|\mathcal{S}_j|}\sum_{\ell\in\mathcal{S}_j}h_\ell^{(s)}.
\end{equation}
When \(\mathcal{A}_i\) contains only its original word, these definitions reduce to that word's token and frame spans. When it also contains attached markers, the word and all markers share one text vector, while their speech vectors follow the corresponding word-internal or inter-word intervals.

Chunk aggregation is applied to both word-level views. Let \(\mathcal{C}=(C_1,\ldots,C_N)\) be an ordered partition of \(U\) into \(N\) chunks. Each sequence of consecutive word units is mean-pooled into one lexical chunk, while each pause, elongation, or boundary unit remains a singleton. For each \(C_k\in\mathcal{C}\), we compute the text and speech representations:

\begin{equation}
c_k^{(t)}=\frac{1}{|C_k|}\sum_{U_j\in C_k}u_j^{(t)},
\quad
c_k^{(s)}=\frac{1}{|C_k|}\sum_{U_j\in C_k}u_j^{(s)}.
\end{equation}
Mean-pooling the \(N\) text and speech chunk representations yields the chunk-pooled features \(x_t\) and \(x_s\), respectively.
In Figure~\ref{fig:unitization-chunking}, the ten units form seven chunks: the consecutive word units in \texttt{The boy} and \texttt{is stealing cookies} are mean-pooled into two lexical chunks, while the four event units and final boundary unit remain singletons. Repeated markers therefore remain separate and each contributes once to the final mean, preserving event magnitude in both modalities; only consecutive word units are pooled.

\subsection{Text-Anchored Paralinguistic Fusion}

The preceding stage yields local text and speech blocks organized around words and prosodic events. Two utterance-level speech blocks complement them with recording-wide vocal characteristics. To account for unequal modality contributions, all four blocks condition a dynamic gate that adjusts the three speech-derived blocks per recording, while text enters as the direct anchor.

From each complete recording, eGeMAPSv02 Functionals extracted with openSMILE \citep{eyben2010opensmile,eyben2016gemaps} yield a handcrafted feature vector \(x_p\) that captures prosody, phonation, and voice quality. In parallel, the pretrained WavLM Base+ encoder \cite{chen2022wavlm} produces frame-level hidden states, which are temporally mean-pooled into the acoustic representation \(x_a\). Exact feature dimensionalities are provided in the implementation details. WavLM complements eGeMAPS by capturing broader speech patterns not explicitly characterized by predefined acoustic descriptors.

The model consequently contains four blocks: \(x_t\), the LLM text representation; \(x_s\), the locally aggregated Whisper speech representation; and \(x_p\) and \(x_a\), the utterance-level eGeMAPS and WavLM blocks, respectively. The mean-pooled \(x_t\) and \(x_s\) first undergo sample-wise LayerNorm \citep{ba2016layernorm} with fixed unit scale and zero bias. All four blocks are then standardized dimension-wise using statistics estimated from the actual training data used in each run. Let \(\hat{x}_q\) denote the resulting block for \(q\in\{t,s,p,a\}\).

Before its block-specific summary projection, each standardized block is rescaled as \(\hat{x}_q/\sqrt{d_q}\), where \(d_q\) is its dimensionality, so that higher-dimensional blocks do not dominate gate generation. The concatenated summaries are then used to predict one scalar logit \(o_q\) for each auxiliary block \(q\in\{s,p,a\}\). Because these logits depend on all four blocks, the sample-specific gates are bounded as follows:

\begin{equation}
g_q = 1 + 0.5\tanh(o_q).
\end{equation}
The final predictor layer is zero-initialized, so every gate starts at 1 and remains within \((0.5,1.5)\). Each auxiliary block therefore retains 50--150\% of its standardized baseline magnitude, while the text block remains unscaled. The text block is kept as the direct anchor, while the three auxiliary blocks are gated as follows:

\begin{equation}
\tilde{x} = [\hat{x}_t; g_s \hat{x}_s; g_p \hat{x}_p; g_a \hat{x}_a].
\end{equation}
This identity-initialized design starts from standardized concatenation and learns sample-dependent modulation of the auxiliary blocks during training. The fused representation \(\tilde{x}\) is classified by a 3-layer MLP trained with cross-entropy.

\section{Experimental Setup}

\subsection{Datasets and Transcription}

We evaluate LAPE on two controlled-access DementiaBank benchmarks built from Cookie Theft picture descriptions: ADReSS, which provides audio and manual transcripts for 108 training and 48 test participants, and ADReSSo, which provides audio only for 166 training and 71 test participants. Each participant contributes one sample. ADReSSo is the primary benchmark because its larger, speech-only setting requires both linguistic and acoustic information to be obtained automatically and better reflects fully automatic use; ADReSS serves as an additional benchmark under a manual-transcript condition. For ADReSSo, Whisper large-v3 \citep{radford2023whisper} generates transcripts and word timestamps, whereas for ADReSS we retain the official transcripts for comparability and align them with the recordings using Qwen3-ForcedAligner-0.6B \citep{shi2026qwen3asr}.

\subsection{Evaluation Protocols}

We evaluate each benchmark's official training partition using participant-level five-fold cross-validation (CV) and leave-one-subject-out (LOSO). For CV, each metric is computed separately on each test fold and then averaged across the five folds. For LOSO, because each test fold contains only one participant, we pool the out-of-sample predictions from all folds and compute each metric once on the pooled predictions; fold-wise computation is not applicable to the complete metric set. Main comparisons report accuracy, macro F1-Score, precision, and recall; later analyses use accuracy (Acc) and area under the receiver operating characteristic curve (ROC-AUC), based on AD probabilities. For compactness, tables abbreviate ROC-AUC as Auc.

\subsection{Baselines}

We use Multi-View Gated GAT (MVG-GAT; \citealp{graphad2026}) and CogniAlign \citep{cognialign2025} as the two cross-setting baselines. MVG-GAT models semantic, syntactic, and discourse-flow structure with adaptive fusion, whereas CogniAlign provides a multimodal reference based on token-level audio--text alignment and cross-attention fusion. To keep Tables~\ref{tab:main-cv} and~\ref{tab:main-loso} compact, we use short row labels rather than in-table citations: FFeat-SVM \citep{robustacoustic2026}, LLaMA2-PL \citep{llmprompt2025}, W2V-BERT \citep{ying2023multimodal}, LLM-A-SVC \citep{llmasvc2025}, Wrap-Ridge \citep{wrapperfs2026}, and GPT-Op \citep{bang2024llm}.

\begin{table*}[t]
\centering
\begingroup
\setlength{\tabcolsep}{4.0pt}
\renewcommand{\arraystretch}{1.05}
\begin{tabular}{L{\methodcolwidth}|*{4}{C{\metriccolwidth}}@{\hspace{1.5mm}\vrule width 1pt\hspace{1.5mm}}L{\methodcolwidth}|*{4}{C{\metriccolwidth}}}
\toprule
ADReSS & Accuracy & F1-Score & Precision & Recall & ADReSSo & Accuracy & F1-Score & Precision & Recall \\
\midrule
FFeat-SVM & 86.90 & 86.90 & 86.90 & 86.90 & W2V-BERT & 83.70 & 83.60 & 85.30 & 84.00 \\
LLaMA2-PL & 81.31 & 81.13 & 82.69 & 79.63 & LLM-A-SVC & \underline{91.80} & \underline{91.80} & \underline{94.60} & 88.20 \\
MVG-GAT & \underline{89.81} & \underline{89.91} & 89.09 & \underline{90.74} & MVG-GAT & 88.81 & 88.81 & 89.43 & 88.81 \\
CogniAlign & 88.73 & 87.38 & \underline{94.00} & 82.91 & CogniAlign & 90.36 & 90.11 & 90.15 & \underline{90.76} \\
\midrule
\textbf{LAPE} & \textbf{98.18} & \textbf{98.18} & \textbf{100.00} & \textbf{96.36} & \textbf{LAPE} & \textbf{94.01} & \textbf{94.01} & \textbf{96.39} & \textbf{92.03} \\
\bottomrule
\end{tabular}
\endgroup
\caption{Cross-validation (CV) results on ADReSS and ADReSSo (\%).}
\label{tab:main-cv}
\end{table*}

\begin{table*}[t]
\centering
\begingroup
\setlength{\tabcolsep}{4.0pt}
\renewcommand{\arraystretch}{1.05}
\begin{tabular}{L{\methodcolwidth}|*{4}{C{\metriccolwidth}}@{\hspace{1.5mm}\vrule width 1pt\hspace{1.5mm}}L{\methodcolwidth}|*{4}{C{\metriccolwidth}}}
\toprule
ADReSS & Accuracy & F1-Score & Precision & Recall & ADReSSo & Accuracy & F1-Score & Precision & Recall \\
\midrule
Wrap-Ridge & 86.50 & 86.60 & 86.08 & 87.18 & GPT-Op & 87.32 & 87.25 & 88.06 & \underline{87.32} \\
MVG-GAT & 87.04 & 86.79 & \underline{88.46} & 85.19 & MVG-GAT & 86.75 & 86.25 & \textbf{94.52} & 79.31 \\
CogniAlign & \underline{87.96} & \underline{88.07} & 87.27 & \underline{88.89} & CogniAlign & \underline{87.35} & \underline{87.57} & 90.24 & 85.06 \\
\midrule
\textbf{LAPE} & \textbf{95.37} & \textbf{95.37} & \textbf{96.23} & \textbf{94.44} & \textbf{LAPE} & \textbf{91.57} & \textbf{91.56} & \underline{92.94} & \textbf{90.80} \\
\bottomrule
\end{tabular}
\endgroup
\caption{Leave-one-subject-out (LOSO) results on ADReSS and ADReSSo (\%).}
\label{tab:main-loso}
\end{table*}

\subsection{Implementation Details}

Unless otherwise stated, LAPE uses the Joint protocol and the chunk-pooled features described above. The four protocols and aggregation levels are analyzed separately below.

Text features are extracted with Kimi-K2.7-Code, which shares the Kimi K2.5 architecture \citep{kimiteam2026k25}, via llama.cpp.\footnote{\url{https://github.com/ggml-org/llama.cpp}.} The four feature blocks \(x_t\), \(x_s\), \(x_p\), and \(x_a\) have dimensions 7168, 768, 88, and 768, respectively, yielding an 8792-D classifier input. NormGate maps each block to a 32-dimensional summary and predicts the three speech-side gates with a dropout-free \(128\rightarrow64\rightarrow3\) MLP; the final task head is a 3-layer MLP with ReLU and dropout.

Training uses AdamW \citep{loshchilov2019adamw} and gradient clipping at 1.0. The early-stopping patience is set to 15. We fix the learning rate, weight decay, dropout, and label smoothing to \(5\mathrm{e}{-}4\), \(1\mathrm{e}{-}4\), \(0.3\), and \(0.1\). In LOSO, \(15\%\) of the remaining training data is reserved for validation. We then use grid search over \(H_1\in\{256,512,768,1024\}\), \(H_2\in\{64,128,256\}\), and batch size \(\in\{8,16,32\}\).

\section{Experimental Results}

\paragraph{Main Results.} Tables~\ref{tab:main-cv} and~\ref{tab:main-loso} report the primary CV and LOSO comparisons. Because LAPE's macro F1-Score differs from accuracy by at most 0.01\%, we focus on accuracy. The small official test sets produce coarse metric increments and split sensitivity; participant-level CV and LOSO instead evaluate every participant out of sample, and agreement between them provides a more stable assessment. LAPE ranks first throughout, reaching 98.18\% \& 94.01\% under CV and 95.37\% \& 91.57\% under LOSO on ADReSS and ADReSSo. Its lead over the strongest listed baseline spans 2.21--8.37\%, averaging 5.55\%.

LOSO reduces LAPE's accuracy by 2.81\% on ADReSS and 2.44\% on ADReSSo, yet its margins remain 7.41\% and 4.22\%, respectively, compared with 8.37\% and 2.21\% under CV. LAPE therefore remains first under the stricter subject-wise evaluation. Its accuracy on ADReSSo is 3.80--4.17\% lower than on ADReSS, consistent with the fully automatic speech-only setting, although dataset differences preclude attributing this gap to transcription alone.

\FloatBarrier

\begin{table*}[t]
\centering
\begingroup
\setlength{\tabcolsep}{2.97pt}
\renewcommand{\arraystretch}{1.05}
\begin{tabular}{L{\transcriptcolwidth}|*{2}{C{\transcriptmetriccolwidth}}|*{2}{C{\transcriptmetriccolwidth}}@{\hspace{1.5mm}\vrule width 1pt\hspace{1.5mm}}L{\transcriptcolwidth}|*{2}{C{\transcriptmetriccolwidth}}|*{2}{C{\transcriptmetriccolwidth}}}
\toprule
ADReSS & \multicolumn{2}{c|}{CV} & \multicolumn{2}{c@{\hspace{1.5mm}\vrule width 1pt\hspace{1.5mm}}}{LOSO} & ADReSSo & \multicolumn{2}{c|}{CV} & \multicolumn{2}{c}{LOSO} \\
\midrule
Protocol & Acc & Auc & Acc & Auc & Protocol & Acc & Auc & Acc & Auc \\
\midrule
Basic    & 92.73 & 95.21 & 89.81 & 93.14 & Basic    & \underline{92.19} & 94.66 & \underline{87.95} & \underline{92.80} \\
Pause    & \underline{96.36} & \textbf{99.01} & \underline{91.67} & \underline{95.85} & Pause    & \underline{92.19} & \textbf{94.86} & 86.75 & 91.23 \\
Elong    & 91.55 & 88.68 & 86.11 & 89.51 & Elong    & 92.17 & 94.01 & 86.75 & 91.91 \\
\midrule
\textbf{Joint} & \textbf{98.18} & \underline{97.19} & \textbf{95.37} & \textbf{96.95}
         & \textbf{Joint} & \textbf{94.01} & \underline{94.74} & \textbf{91.57} & \textbf{93.70} \\
\bottomrule
\end{tabular}
\endgroup
\caption{Comparison of transcription protocols on ADReSS and ADReSSo (\%).}
\label{tab:transcript}

\begingroup
\setlength{\tabcolsep}{2.97pt}
\renewcommand{\arraystretch}{0.95}
\begin{tabular}{L{\transcriptcolwidth}|*{2}{C{\transcriptmetriccolwidth}}|*{2}{C{\transcriptmetriccolwidth}}@{\hspace{1.5mm}\vrule width 1pt\hspace{1.5mm}}L{\transcriptcolwidth}|*{2}{C{\transcriptmetriccolwidth}}|*{2}{C{\transcriptmetriccolwidth}}}
\toprule
ADReSS & \multicolumn{2}{c|}{CV} & \multicolumn{2}{c@{\hspace{1.5mm}\vrule width 1pt\hspace{1.5mm}}}{LOSO} & ADReSSo & \multicolumn{2}{c|}{CV} & \multicolumn{2}{c}{LOSO} \\
\midrule
Level & Acc & Auc & Acc & Auc & Level & Acc & Auc & Acc & Auc \\
\midrule
Token & \underline{95.45} & \textbf{97.19} & 89.81 & \underline{96.33} & Token & \underline{93.33} & \underline{94.06} & 88.55 & 93.05 \\
Word  & \underline{95.45} & \underline{97.02} & \underline{92.59} & 94.44 & Word  & 92.82 & 93.95 & \underline{89.16} & \underline{93.53} \\
\midrule
\textbf{Chunk} & \textbf{98.18} & \textbf{97.19} & \textbf{95.37} & \textbf{96.95} & \textbf{Chunk} & \textbf{94.01} & \textbf{94.74} & \textbf{91.57} & \textbf{93.70} \\
\bottomrule
\end{tabular}
\endgroup
\caption{Aggregation comparison on ADReSS and ADReSSo (\%).}
\label{tab:granularity}
\end{table*}

\begin{table*}[t]
\centering
\begingroup
\normalsize
\setlength{\tabcolsep}{2.5pt}
\renewcommand{\arraystretch}{1.05}
\begin{tabular}{@{}L{\ablationvariantcolwidth}|*{2}{C{\ablationmetriccolwidth}}|*{2}{C{\ablationmetriccolwidth}}@{\hspace{1.5mm}\vrule width 1pt\hspace{1.5mm}}L{\ablationvariantcolwidth}|*{2}{C{\ablationmetriccolwidth}}|*{2}{C{\ablationmetriccolwidth}}@{}}
\toprule
ADReSS & \multicolumn{2}{c|}{CV} & \multicolumn{2}{c@{\hspace{1.5mm}\vrule width 1pt\hspace{1.5mm}}}{LOSO} & ADReSSo & \multicolumn{2}{c|}{CV} & \multicolumn{2}{c}{LOSO} \\
\midrule
Variant & Acc & Auc & Acc & Auc & Variant & Acc & Auc & Acc & Auc \\
\midrule
Text only      & 96.36 & 96.20 & 92.59 & 94.51 & Text only      & 92.80 & \underline{94.49} & 88.55 & \textbf{93.92} \\
Speech only    & 87.09 & 86.86 & 80.56 & 85.19 & Speech only    & 89.21 & 90.58 & 81.33 & 86.21 \\
Local Feats.   & \underline{97.27} & \textbf{98.35} & \textbf{95.37} & \underline{96.43} & Local Feats.   & \underline{93.40} & 94.07 & 88.55 & 92.07 \\
w/o NG        & \underline{97.27} & \underline{97.52} & 92.59 & 95.78 & w/o NG        & 93.39 & 94.05 & \underline{90.36} & 92.29 \\
\midrule
\textbf{LAPE}  & \textbf{98.18} & 97.19 & \textbf{95.37} & \textbf{96.95} & \textbf{LAPE} & \textbf{94.01} & \textbf{94.74} & \textbf{91.57} & \underline{93.70} \\
\bottomrule
\end{tabular}
\endgroup
\caption{Component ablation on ADReSS and ADReSSo under CV and LOSO (\%). NG abbreviates NormGate; w/o NG is the variant without it. The local-feature variant uses only the chunk-pooled text and Whisper speech representations, with NormGate disabled.}
\label{tab:ablation}
\end{table*}

\paragraph{Transcription Protocol Analysis.} Table~\ref{tab:transcript} compares the four protocols with Kimi and chunk aggregation under the same search grid and budget. Relative to Basic, Pause improves ADReSS by 3.63\% under CV and 1.86\% under LOSO, but changes ADReSSo by 0.00\% and \ensuremath{-}1.20\%; Elong changes the four settings by \ensuremath{-}1.18\%, \ensuremath{-}3.70\%, \ensuremath{-}0.02\%, and \ensuremath{-}1.20\%. Thus, neither single-event protocol is reliable alone. Joint instead improves over Basic by 5.45\%, 5.56\%, 1.82\%, and 3.62\%, averaging 4.11\%, and exceeds the better single-event protocol by 2.74\% on average. Its gain is larger under LOSO than CV on both datasets, especially ADReSSo (3.62\% vs. 1.82\%). This supports complementary roles: pauses mark interruptions, whereas elongations capture prolonged speech in context, making Joint more robust to held-out speakers.

\paragraph{Aggregation Analysis.} Table~\ref{tab:granularity} compares three granularities under Joint: Token retains subword elements, Word uses lexico-prosodic units, and Chunk mean-pools consecutive word units while keeping each pause, elongation, and boundary unit separate. All other settings are unchanged.

Moving from Token to Word is inconsistent, changing accuracy by 0.00\%, 2.78\%, \ensuremath{-}0.51\%, and 0.61\% across ADReSS CV, ADReSS LOSO, ADReSSo CV, and ADReSSo LOSO. Chunk ranks first throughout and exceeds the better fine-grained alternative by 2.73\%, 2.78\%, 0.68\%, and 2.41\% (2.15\% on average). Relative to Token, its mean gain grows from 1.71\% under CV to 4.29\% under LOSO, suggesting that pooling consecutive word units suppresses speaker-specific redundancy while preserving sparse paralinguistic events. Together, Tables~\ref{tab:transcript} and~\ref{tab:granularity} show that complementary event markers must remain salient rather than being diluted across a long token sequence.

\paragraph{Component Ablation.} Table~\ref{tab:ablation} first contrasts the two unimodal variants: text only uses the text anchor, whereas speech only combines local Whisper with both utterance-level speech blocks. It then follows a nested multimodal path, adding local Whisper to text, the utterance-level blocks without NormGate, and finally full LAPE. Because NormGate includes normalization and bounded gating, the last transition measures their joint contribution.

Text only outperforms speech only by 9.27\% \& 12.03\% on ADReSS and 3.59\% \& 7.22\% on ADReSSo under CV and LOSO. The larger LOSO gaps on both datasets suggest greater speaker sensitivity in speech representations and support the text anchor. Adding local Whisper never reduces accuracy, yielding gains of 0.91\%, 2.78\%, 0.60\%, and 0.00\% (1.07\% on average). These consistent gains show that local acoustics complement linguistic content, with the largest benefit under ADReSS LOSO.

By contrast, utterance-level blocks without NormGate help only ADReSSo LOSO and are neutral or worse elsewhere. NormGate raises accuracy by 0.91\%, 2.78\%, 0.62\%, and 1.21\% over this variant (1.38\% on average). Its larger mean gain under LOSO than CV (2.00\% vs. 0.77\%) indicates greater benefit for held-out speakers. Auc also rises in three of four settings, except ADReSS CV, supporting adaptive scaling of global speech cues against the text anchor.

\FloatBarrier

\section{Conclusion}

We presented LAPE, an LLM-anchored paralinguistic enrichment framework for speech-based AD detection. LAPE enriches LLM-derived linguistic representations through prosodic event textualization, lexico-prosodic unitization and chunking, and text-anchored fusion of local and utterance-level speech features. It achieves state-of-the-art accuracy and F1 on ADReSS and ADReSSo under both CV and LOSO. Future work will extend LAPE toward speech-aware reasoning by textualizing richer paralinguistic cues and distilling this capability into smaller LLMs without sacrificing diagnostic performance.

\section{Generative AI Use Disclosure}
ChatGPT was used solely to polish text originally written by the authors. All suggestions were reviewed and revised by the authors, who take full responsibility for the manuscript. The tool was not used to generate scientific content or references.

\end{document}


\maketitle

This supplement reports implementation details, data provenance, and descriptive
results omitted from the main paper.  Definitions and equations given there are
not repeated.  Unless noted otherwise, descriptive checks use the final ADReSS
and ADReSSo feature inputs, the Joint protocol (pause and elongation markers),
and event-preserving chunk aggregation.

\section{Timestamp and Marker Checks}

The main paper defines the pause and elongation equations and final bins; this
section documents the syllable heuristic, reference anchors, and timestamp
filtering.

\begin{lstlisting}
estimate_syllables(word):
    cleaned = lowercase(word), then delete characters outside [a-z]
    n = number of contiguous [aeiouy]+ groups
    if cleaned ends with e, not le, and n > 1: n -= 1
    if cleaned ends with ed, not [td]ed, and n > 1: n -= 1
    return max(1, n)
\end{lstlisting}

Nonalphabetic characters are removed; no dictionary or upper bound is used, and
empty or vowel-free strings return one syllable.
This proxy is used only for duration normalization; it does not modify the
transcript text or tokenizer inputs.

Anchors were derived once from all 237 ADReSSo word-timestamp files (166
official training and 71 official test-distribution files).  Of 24,676 words,
filtering removed 71 empty or filler items, 884 outside
the $(40,2500]$ ms duration range, and 221 with mean duration per estimated
syllable above 1,500 ms, leaving 23,500 words.  Table~\ref{tab:duration-anchor}
summarizes the resulting statistics and final smoothing.
The median column gives the empirical reference, whereas the final-anchor
column lists the values used in the binning rule, including the smoothed
three-syllable value and the extrapolation for five or more syllables.

These checks make preprocessing auditable: syllable estimates set durations,
anchors set residuals, and tag counts define events.

\newpage
\noindent\begin{minipage}{\columnwidth}
We count a contiguous run of one to three identical tags as one event while
retaining repeated tags as separate marker instances; removing all markers
recovers the corresponding plain transcript.  This distinction separates event
frequency from the number of inserted markers and supports the input-expansion
statistics reported below.
\end{minipage}

\begin{table}[t]
\centering
\tablecompact
\begin{tabular*}{\columnwidth}{@{\extracolsep{\fill}}clrc@{}}
\toprule
Syllables & Words & Median & Final anchor \\
\midrule
1\phantom{+} & 18,141 & 260 ms & 260 ms \\
2\phantom{+} & 4,750  & 360 ms & 360 ms \\
3\phantom{+} & 457    & 420 ms & 450 ms \\
4\phantom{+} & 145    & 580 ms & 580 ms \\
5+ & 7     & 760 ms & $700+150(n-5)$ ms \\
\bottomrule
\end{tabular*}
\caption{Duration-anchor provenance after filtering.  The final mathematical
definition is given in the main paper.}
\label{tab:duration-anchor}
\end{table}

\begin{table}[t]
\centering
\tablecompact
\setlength{\tabcolsep}{0pt}
\begin{tabular*}{\columnwidth}{@{\extracolsep{\fill}}L{0.34\columnwidth}|*{2}{C{0.22\columnwidth}}@{}}
\toprule
Statistic & ADReSS & ADReSSo \\
\midrule
With Pause (\%) & 100.00 & 100.00 \\
Pause events & 14.96 & 8.86 \\
\midrule
With Elong. (\%) & 97.22 & 99.40 \\
Elong. events & 10.81 & 20.28 \\
\midrule
All marker events & 25.78 & 29.13 \\
Markers per event & 1.63 & 1.77 \\
\midrule
Token number & 232.68 & 284.10 \\
Unit number & 147.70 & 175.51 \\
Chunk number & 65.82 & 106.26 \\
\bottomrule
\end{tabular*}
\caption{Marker coverage and input expansion.  Percentages are participant
proportions; count cells are mean counts per participant except for markers per
event.}
\label{tab:marker-coverage}
\end{table}

Table~\ref{tab:marker-coverage} shows that pause information is present for
every participant, while elongation information is present for nearly every
participant.  ADReSS has more pause but fewer elongation events, so its total
marker-event count is lower (25.78 versus 29.13); ADReSSo has larger token,
unit, and chunk counts.  Values above one marker per event reflect repeated-tag
encoding of longer bins, not duplicate events.  Chunking reduces 147.70 units to
65.82 chunks for ADReSS and 175.51 to 106.26 for ADReSSo.  These are descriptive
input properties, not performance comparisons.

\begin{table}[t]
\centering
\tablecompact
\setlength{\tabcolsep}{3pt}
\begin{tabular*}{\columnwidth}{@{\extracolsep{\fill}}lccc@{}}
\toprule
Bin & Threshold & ADReSS & ADReSSo \\
\midrule
P1 & $(500,1000]$ & \phantom{0}592 (36.63\%) & \phantom{0}289 (19.66\%) \\
P2 & $(1000,2000]$ & \phantom{0}536 (33.17\%) & \phantom{0}443 (30.14\%) \\
P3 & $(2000,\infty)$ & \phantom{0}488 (30.20\%) & \phantom{0}738 (50.20\%) \\
\midrule
E1 & $(200,500]$ & \phantom{0}994 (85.10\%) & 2,240 (66.55\%) \\
E2 & $(500,800]$ & \phantom{0}110 (9.42\%) & \phantom{0}470 (13.96\%) \\
E3 & $(800,\infty)$ & \phantom{00}64 (5.48\%) & \phantom{0}656 (19.49\%) \\
\bottomrule
\end{tabular*}
\caption{Distribution of duration-bin events; thresholds are in milliseconds
(ms).  Percentages use all events of the same type within each dataset.  P1--P3
denote pause bins and E1--E3 elongation bins, whose thresholds are residual
durations beyond the expected syllable-based word duration.}
\label{tab:marker-bins}
\end{table}

\section{Prompt Design and Serialization}

The prompt combines a shared task instruction with a marker note inside
\texttt{[INSTRUCTION]}; the former fixes target/output and the latter defines
markers/bins.  This separation lets Basic omit marker context.

\subsection{Task Instruction}

Five instruction variants were evaluated; no stable ordering emerged, so the
canonical wording is:

\noindent\begin{minipage}{\columnwidth}
\begin{lstlisting}[breakatwhitespace=true,breakindent=1em,aboveskip=7pt,belowskip=7pt]
Classify the following transcript into one of two labels:
AD = Alzheimer's disease patient
HC = Healthy / non-AD control
Focus on cognitive-linguistic signals in the transcript.
Output AD or HC only.
\end{lstlisting}
\end{minipage}

Adding task-instruction context produced positive descriptive differences for
all four transcript views in a development-time ADReSSo check.  The check used
word-level text-only features, the 108-configuration search space,
accuracy-based early stopping, and mean validation Macro-F1 for trial
selection; each condition was independently tuned, including the seed.  It is
therefore not a controlled ablation.  Table~\ref{tab:prompt-deltas} is an
exception to the Joint default: it reports a descriptive sensitivity check over
complete instruction-conditioned inputs, not a task-instruction-only ablation.

\subsection{Transcript Instruction}

This note follows the task instruction and defines the marker
vocabulary and duration bins.  The exact ADReSSo wording is:

\begin{lstlisting}[breakatwhitespace=true,breakindent=1em,aboveskip=7pt,belowskip=7pt]
Transcript note: In these transcripts, <pause> tags encode binned pause duration (one: >500ms and <=1000ms; two: >1000ms and <=2000ms; three: >2000ms), and <elongated> tags appended to words encode binned extra duration beyond the expected syllable-based word duration (one: >200ms and <=500ms extra; two: >500ms and <=800ms extra; three: >800ms extra).
\end{lstlisting}

\newpage
\begin{table}[t]
\centering
\normalsize
\setlength{\tabcolsep}{0pt}
\renewcommand{\arraystretch}{1.02}
\begin{tabular*}{\columnwidth}{@{\extracolsep{\fill}}L{0.22\columnwidth}|*{5}{C{0.15\columnwidth}}@{}}
\toprule
Protocol & $\Delta$ Acc. & $\Delta$ F1 & $\Delta$ Pre. & $\Delta$ Rec. & $\Delta$ Auc \\
\midrule
Basic & +1.82 & +1.87 & +0.69 & +2.16 & +1.91 \\
Pause & +3.05 & +3.09 & +2.00 & +3.60 & +2.84 \\
Elong. & +1.17 & +1.20 & +3.93 & -2.22 & +1.79 \\
\midrule
Joint & +1.19 & +1.23 & +0.90 & +1.04 & -0.56 \\
\bottomrule
\end{tabular*}
\caption{Effect of the task instruction on ADReSSo word-level text-only
classification, by transcript protocol.  Each row reports the
instruction-present minus instruction-absent difference (\%) for one
independently tuned protocol.}
\label{tab:prompt-deltas}
\end{table}

Pause has the largest gains in F1 (+3.09) and recall (+3.60), whereas Elongation
trades a precision gain (+3.93) for lower recall (-2.22).  Joint is smaller and
slightly lowers Auc, so these values remain a sensitivity check rather than a
claim of uniform superiority.

\pagebreak[3]
\begin{samepage}
\subsection{Serialization and Token States}

The marker-enhanced request layout is:

\noindent\begin{minipage}{\columnwidth}
\begin{lstlisting}[aboveskip=7pt]
[INSTRUCTION]
{task instruction}

Transcript note: {transcript instruction body}

[TRANSCRIPT]
{enhanced transcript}
\end{lstlisting}
\end{minipage}

All variants include the task instruction; only Pause, Elongation, and Joint
include the transcript note, while Basic has no note because it contains no
timing markers.

The strings \pausemark{} and \elongmark{} are ordinary literals, not tokenizer
special tokens.  Both datasets use no demonstrations and mean-pool retained
subword states; character spans select transcript-overlapping rows, excluding
instruction rows from pooling while retaining contextualization from the full
prompt.  The llama.cpp context limit was 1,536 tokens; the longest requests
used 958 tokens for ADReSS and 1,103 for ADReSSo.

\paragraph{Timestamp provenance.}
Timestamp generation is dataset-specific and precedes all encoding experiments.
ADReSSo uses \texttt{openai-whisper==20250625} with \texttt{large-v3}; complete
recordings are transcribed, INV-overlapping words are removed, and the same 237
word-timestamp CSV files are reused on the original recording timeline.  ADReSS
keeps the manual CHAT participant transcript as authoritative and uses
\texttt{Qwen/Qwen3-ForcedAligner-0.6B} through \texttt{qwen-asr==0.0.6} for
timestamps only.
\end{samepage}

\section{Experimental Setup and Reproduction}

\paragraph{Computing environment.}

Experiments ran on 16 NVIDIA GeForce RTX 4090 D GPUs (24 GB each) across two
nodes.  Each node had two Intel Xeon Gold 6133 CPUs and approximately 256 GB
RAM.  The audited stack was Ubuntu 22.04.5, driver 595.84, Python 3.11.15,
PyTorch 2.8.0/CUDA 12.8, Transformers 4.57.6, OpenSMILE 2.6.0,
openai-whisper 20250625, and qwen-asr 0.0.6; all remaining dependencies were
pinned in the release specification.

\begin{table*}[!t]
\centering
\begingroup
\normalsize
\setlength{\tabcolsep}{1.5pt}
\renewcommand{\arraystretch}{1.10}
\begin{tabular*}{\textwidth}{@{\extracolsep{\fill}}L{\expandedvariantcolwidth}@{\hspace{1.5mm}\vrule width 1pt\hspace{1.5mm}}*{4}{C{\expandedcvmetriccolwidth}}@{\hspace{1.5mm}\vrule width 1pt\hspace{1.5mm}}*{4}{C{\expandedcvmetriccolwidth}}@{}}
\toprule
CV & \multicolumn{4}{c@{\hspace{1.5mm}\vrule width 1pt\hspace{1.5mm}}}{ADReSS} & \multicolumn{4}{c}{ADReSSo} \\
\midrule
Variant & Acc. & F1 & Pre. & Rec. & Acc. & F1 & Pre. & Rec. \\
\midrule
Basic & \msd{92.73}{7.39} & \msd{92.64}{7.47} & \msd{97.78}{4.44} & \msd{87.27}{12.33} & \runnerup{\msd{92.19}{3.96}} & \runnerup{\msd{92.19}{3.96}} & \msd{95.22}{4.42} & \msd{89.67}{4.17} \\
Pause & \runnerup{\msd{96.36}{4.45}} & \runnerup{\msd{96.35}{4.47}} & \runnerup{\msd{98.18}{3.64}} & \runnerup{\msd{94.55}{7.27}} & \runnerup{\msd{92.19}{3.96}} & \msd{92.17}{3.96} & \msd{93.16}{3.99} & \best{\msd{92.09}{7.50}} \\
Elong. & \msd{91.55}{5.60} & \msd{91.45}{5.69} & \msd{96.33}{4.52} & \msd{86.73}{11.66} & \msd{92.17}{3.00} & \msd{92.16}{3.01} & \best{\msd{96.53}{4.52}} & \msd{88.56}{6.03} \\
\midrule
Token & \runnerup{\msd{95.45}{4.07}} & \msd{95.44}{4.08} & \runnerup{\msd{98.18}{3.64}} & \runnerup{\msd{92.73}{6.80}} & \runnerup{\msd{93.33}{3.64}} & \runnerup{\msd{93.30}{3.66}} & \runnerup{\msd{95.54}{5.47}} & \runnerup{\msd{91.90}{2.99}} \\
Word & \runnerup{\msd{95.45}{4.98}} & \runnerup{\msd{95.45}{4.99}} & \msd{98.00}{4.00} & \runnerup{\msd{92.73}{6.80}} & \msd{92.82}{6.76} & \msd{92.81}{6.77} & \msd{94.11}{6.45} & \best{\msd{92.03}{7.77}} \\
\midrule
Text only & \msd{96.36}{3.40} & \msd{96.34}{3.43} & \runnerup{\msd{98.33}{3.33}} & \runnerup{\msd{94.55}{7.27}} & \msd{92.80}{5.48} & \msd{92.79}{5.49} & \msd{95.14}{4.45} & \msd{90.78}{6.84} \\
Speech only & \msd{87.09}{4.32} & \msd{87.06}{4.36} & \msd{88.51}{3.43} & \msd{85.27}{7.14} & \msd{89.21}{2.88} & \msd{89.14}{2.90} & \msd{89.19}{4.77} & \runnerup{\msd{90.98}{6.60}} \\
Local Feats. & \runnerup{\msd{97.27}{2.23}} & \runnerup{\msd{97.27}{2.23}} & \runnerup{\msd{98.33}{3.33}} & \best{\msd{96.36}{4.45}} & \runnerup{\msd{93.40}{3.42}} & \runnerup{\msd{93.40}{3.42}} & \runnerup{\msd{96.32}{3.00}} & \msd{90.85}{4.42} \\
w/o NG & \runnerup{\msd{97.27}{3.64}} & \runnerup{\msd{97.27}{3.64}} & \msd{98.18}{3.64} & \best{\msd{96.36}{4.45}} & \msd{93.39}{3.50} & \msd{93.38}{3.50} & \msd{96.31}{3.02} & \msd{90.78}{4.70} \\
\midrule
\textbf{LAPE} & \best{\msd{98.18}{2.23}} & \best{\msd{98.18}{2.23}} & \best{\msd{100.00}{0.00}} & \best{\msd{96.36}{4.45}} & \best{\msd{94.01}{3.72}} & \best{\msd{94.01}{3.72}} & \best{\msd{96.39}{2.96}} & \best{\msd{92.03}{5.72}} \\
\bottomrule
\end{tabular*}
\endgroup
\caption{Five-fold CV metrics (\%) for transcription-protocol comparisons at
chunk level, aggregation-level comparisons under Joint transcription, and
component ablations under Joint transcription with chunk aggregation; the
shared Joint/Chunk/LAPE configuration is reported once as the final LAPE row.
Each cell is mean $\pm$ fold SD.}
\label{tab:cv-expanded}
\label{tab:transcription-cv-expanded}
\label{tab:aggregation-cv-expanded}
\label{tab:ablation-cv-expanded}
\vspace{\dblfloatsep}
\centering
\begingroup
\normalsize
\setlength{\tabcolsep}{1.5pt}
\renewcommand{\arraystretch}{1.10}
\begin{tabular*}{\textwidth}{@{\extracolsep{\fill}}L{\expandedvariantcolwidth}@{\hspace{1.5mm}\vrule width 1pt\hspace{1.5mm}}*{5}{C{\expandedlosometriccolwidth}}@{\hspace{1.5mm}\vrule width 1pt\hspace{1.5mm}}*{5}{C{\expandedlosometriccolwidth}}@{}}
\toprule
LOSO & \multicolumn{5}{c@{\hspace{1.5mm}\vrule width 1pt\hspace{1.5mm}}}{ADReSS} & \multicolumn{5}{c}{ADReSSo} \\
\midrule
Variant & Acc. & F1 & Pre. & Rec. & Auc & Acc. & F1 & Pre. & Rec. & Auc \\
\midrule
Basic & 89.81 & 89.81 & 92.16 & 87.04 & 93.14 & \runnerup{87.95} & \runnerup{87.94} & 89.41 & \runnerup{87.36} & \runnerup{92.80} \\
Pause & \runnerup{91.67} & \runnerup{91.67} & \runnerup{92.45} & \runnerup{90.74} & \runnerup{95.85} & 86.75 & 86.75 & \runnerup{90.12} & 83.91 & 91.23 \\
Elong. & 86.11 & 86.11 & 85.45 & 87.04 & 89.51 & 86.75 & 86.75 & \runnerup{90.12} & 83.91 & 91.91 \\
\midrule
Token & 89.81 & 89.81 & 92.16 & 87.04 & \runnerup{96.33} & 88.55 & 88.55 & \runnerup{92.50} & 85.06 & 93.05 \\
Word & \runnerup{92.59} & \runnerup{92.59} & \runnerup{94.23} & \runnerup{90.74} & 94.44 & \runnerup{89.16} & \runnerup{89.15} & 91.57 & \runnerup{87.36} & \runnerup{93.53} \\
\midrule
Text only & \runnerup{92.59} & 92.59 & 94.23 & 90.74 & 94.51 & 88.55 & 88.55 & \runnerup{91.46} & \runnerup{86.21} & \best{93.92} \\
Speech only & 80.56 & 80.42 & 86.67 & 72.22 & 85.19 & 81.33 & 81.31 & 83.33 & 80.46 & 86.21 \\
Local Feats. & \best{95.37} & \runnerup{95.36} & \best{100.00} & 90.74 & \runnerup{96.43} & 88.55 & 88.55 & \runnerup{91.46} & \runnerup{86.21} & 92.07 \\
w/o NG & \runnerup{92.59} & 92.59 & 92.59 & \runnerup{92.59} & 95.78 & \runnerup{90.36} & \runnerup{90.34} & 90.80 & \best{90.80} & 92.29 \\
\midrule
\textbf{LAPE} & \best{95.37} & \best{95.37} & \runnerup{96.23} & \best{94.44} & \best{96.95} & \best{91.57} & \best{91.56} & \best{92.94} & \best{90.80} & \runnerup{93.70} \\
\bottomrule
\end{tabular*}
\endgroup
\caption{Pooled LOSO metrics (\%) for the same three analyses under the fixed
conditions defined in Table~\ref{tab:cv-expanded}.}
\label{tab:loso-expanded}
\label{tab:transcription-loso-expanded}
\label{tab:aggregation-loso-expanded}
\label{tab:ablation-loso-expanded}
\end{table*}

\FloatBarrier

\paragraph{Data access and reproduction.}

ADReSS and ADReSSo are controlled-access DementiaBank resources requiring
registration, approval, and compliance with the data-use terms.  They are
established, widely used challenge benchmarks for AD detection within the
DementiaBank ecosystem.  Raw participant data and derived features are not
redistributed; the release contains code, configurations, path schemas, and
synthetic fixtures only.

Reproduction validates the authorized manifest and transcript source, generates
or loads timestamps, and serializes the enhanced transcript.  It extracts
instruction-conditioned Kimi states and frozen Whisper rows, constructs
lexical and event units, event-preserving chunks, and utterance-level speech
blocks.  It then records fallbacks, creates participant-level CV or LOSO splits,
fits standardization on the training partition only, and retains prediction
artifacts, configurations, and software versions for table regeneration.

\section{Expanded Main-Paper Metrics}

Tables S5--S6 report the main-paper metrics together with CV fold SD.
With near-balanced classes, macro-F1 closely tracks accuracy (maximum gap:
0.14\%).  Fold variation is concentrated
in recall: ADReSS Basic and Elong. have recall SDs of 12.33 and 11.66
(respectively), versus 0.00 \& 4.45 on ADReSS and 2.96 \& 5.72 on ADReSSo for
LAPE, so the remaining split sensitivity mainly concerns missed AD cases.

Class-specific metrics clarify the protocol results.  On ADReSS, Pause over Basic
raises recall by 7.28\% under CV and 3.70\% under LOSO, but precision by only
0.40\% and 0.29\% (respectively), so its gains mainly reflect fewer misses.  On
ADReSSo CV, equal accuracy hides a trade-off: recall rises by 2.42\% and
precision falls by 2.06\%; under LOSO, recall instead falls by 3.45\%.
Single-marker effects therefore change error type across datasets and
evaluations.

Under LOSO, Chunk over Word raises precision \& recall by 2.00\% \& 3.70\% on ADReSS
and 1.37\% \& 3.44\% on ADReSSo, so its gain is not one-sided.  NormGate over w/o NG
never lowers either measure: precision \& recall changes are +3.64\% \& +1.85\%
and +2.14\% \& +0.00\% under LOSO, and +1.82\% \& +0.00\% and +0.08\% \& +1.25\%
under CV (ADReSS \& ADReSSo, respectively).  Fusion improves error balance
without a precision-recall trade-off.

\begin{table*}[!t]
\centering
\begingroup
\normalsize
\setlength{\tabcolsep}{1.5pt}
\renewcommand{\arraystretch}{1.10}
\begin{tabular*}{\textwidth}{L{1.05cm}L{1.55cm}@{\extracolsep{\fill}}@{\hspace{1.5mm}\vrule width 1pt\hspace{1.5mm}}*{5}{C{\llmmetriccolwidth}}@{\hspace{1.5mm}\vrule width 1pt\hspace{1.5mm}}*{5}{C{\llmmetriccolwidth}}@{}}
\toprule
\multicolumn{2}{c@{\hspace{1.5mm}\vrule width 1pt\hspace{1.5mm}}}{CV} & \multicolumn{5}{c@{\hspace{1.5mm}\vrule width 1pt\hspace{1.5mm}}}{ADReSS} & \multicolumn{5}{c}{ADReSSo} \\
\midrule
LLM & Setting & Acc. & F1 & Pre. & Rec. & Auc & Acc. & F1 & Pre. & Rec. & Auc \\
\midrule
\multirow{2}{*}{MMax} & Plain & 91.73 & 91.62 & 96.36 & 87.27 & \runnerup{95.21} & \runnerup{89.80} & \runnerup{89.78} & \runnerup{92.82} & \runnerup{87.52} & \runnerup{91.91} \\
 & Enriched & \runnerup{93.55} & \runnerup{93.52} & \runnerup{98.00} & \runnerup{88.91} & 94.28 & 89.18 & 89.15 & 91.72 & 87.39 & 91.60 \\
\midrule
\multirow{2}{*}{GLM} & Plain & 93.45 & 93.42 & 95.00 & \runnerup{92.36} & 94.92 & 90.41 & 90.36 & 93.25 & 88.63 & 91.76 \\
 & Enriched & \runnerup{95.45} & \runnerup{95.43} & \best{\runnerup{100.00}} & 90.91 & \runnerup{96.86} & \runnerup{92.21} & \runnerup{92.19} & \runnerup{95.42} & \runnerup{89.74} & \best{\runnerup{94.75}} \\
\midrule
\multirow{2}{*}{Kimi} & Plain & 93.64 & 93.63 & 93.18 & 94.55 & 96.20 & 92.21 & 92.19 & 96.21 & 88.63 & 94.14 \\
  & Enriched & \best{\runnerup{98.18}} & \best{\runnerup{98.18}} & \best{\runnerup{100.00}} & \best{\runnerup{96.36}} & \best{\runnerup{97.19}} & \best{\runnerup{94.01}} & \best{\runnerup{94.01}} & \best{\runnerup{96.39}} & \best{\runnerup{92.03}} & \runnerup{94.74} \\
\bottomrule
\end{tabular*}
\endgroup
\caption{Five-fold CV metrics (\%) for MiniMax, GLM, and Kimi under the same
search grid and budget as the main paper.  \emph{Plain} is Basic protocol with
token aggregation; \emph{Enriched} is Joint protocol with chunk aggregation.
Both settings keep the full fusion stack (local and utterance-level speech
blocks with NormGate); only the transcription protocol and aggregation level
differ.}
\label{tab:llm-backbone-cv}
\vspace{\dblfloatsep}
\centering
\begingroup
\normalsize
\setlength{\tabcolsep}{1.5pt}
\renewcommand{\arraystretch}{1.10}
\begin{tabular*}{\textwidth}{L{1.05cm}L{1.55cm}@{\extracolsep{\fill}}@{\hspace{1.5mm}\vrule width 1pt\hspace{1.5mm}}*{5}{C{\llmmetriccolwidth}}@{\hspace{1.5mm}\vrule width 1pt\hspace{1.5mm}}*{5}{C{\llmmetriccolwidth}}@{}}
\toprule
\multicolumn{2}{c@{\hspace{1.5mm}\vrule width 1pt\hspace{1.5mm}}}{LOSO} & \multicolumn{5}{c@{\hspace{1.5mm}\vrule width 1pt\hspace{1.5mm}}}{ADReSS} & \multicolumn{5}{c}{ADReSSo} \\
\midrule
LLM & Setting & Acc. & F1 & Pre. & Rec. & Auc & Acc. & F1 & Pre. & Rec. & Auc \\
\midrule
\multirow{2}{*}{MMax} & Plain & 87.96 & 87.94 & \runnerup{91.84} & 83.33 & 92.01 & 83.73 & 83.71 & 84.88 & 83.91 & \runnerup{89.41} \\
 & Enriched & \runnerup{88.89} & \runnerup{88.89} & 90.38 & \runnerup{87.04} & \runnerup{93.76} & \runnerup{85.54} & \runnerup{85.49} & \runnerup{85.39} & \runnerup{87.36} & 88.81 \\
\midrule
\multirow{2}{*}{GLM} & Plain & 88.89 & 88.89 & 88.89 & 88.89 & 92.18 & 86.14 & 86.14 & 89.02 & \runnerup{83.91} & 90.43 \\
 & Enriched & \runnerup{93.52} & \runnerup{93.51} & \runnerup{96.08} & \runnerup{90.74} & \runnerup{95.85} & \runnerup{86.75} & \runnerup{86.75} & \runnerup{91.14} & 82.76 & \runnerup{91.12} \\
\midrule
\multirow{2}{*}{Kimi} & Plain & 89.81 & 89.81 & 90.57 & 88.89 & 92.46 & 87.95 & 87.95 & 91.36 & 85.06 & 93.25 \\
  & Enriched & \best{\runnerup{95.37}} & \best{\runnerup{95.37}} & \best{\runnerup{96.23}} & \best{\runnerup{94.44}} & \best{\runnerup{96.95}} & \best{\runnerup{91.57}} & \best{\runnerup{91.56}} & \best{\runnerup{92.94}} & \best{\runnerup{90.80}} & \best{\runnerup{93.70}} \\
\bottomrule
\end{tabular*}
\endgroup
\caption{Pooled LOSO metrics (\%) for the same LLM comparison under the fixed
conditions defined in Table~\ref{tab:llm-backbone-cv}.}
\label{tab:llm-backbone-loso}
\label{tab:llm-backbone}
\end{table*}

\section{LLM Backbone Comparison}

The main experiments use instruction-conditioned Kimi states.  To check whether
the first two LAPE design choices---prosodic event textualization (Basic vs.\
Joint) and event-preserving chunk aggregation (Token vs.\ Chunk)---transfer
across backbones, we compare \emph{Plain} and \emph{Enriched} settings on three
MoE instruction-tuned LLMs covering distinct total and per-token activated
scales: MiniMax-M3 (UD-Q5\_K\_XL; $\approx$428B total / 23B activated), GLM-5.2
(UD-IQ4\_XS; $\approx$744B total / 40B activated), and Kimi-K2.7-Code
(UD-Q2\_K\_XL; $\approx$1T total / 32B activated), all served with llama.cpp.
Each backbone--setting pair is tuned independently under the same search grid
and budget as the main paper;
Tables~\ref{tab:llm-backbone-cv}--\ref{tab:llm-backbone-loso} report the
resulting CV and pooled-LOSO metrics, respectively.  Both settings retain
the full fusion stack (local and utterance-level speech blocks with NormGate);
only the transcription protocol and aggregation level differ.

\paragraph{Quantized footprint.}
Although the three backbones differ widely in total parameters, their usable
quantized weights are close in size: MiniMax-M3 (UD-Q5\_K\_XL) $\approx$300
GiB, Kimi-K2.7-Code (UD-Q2\_K\_XL) $\approx$320 GiB, and GLM-5.2
(UD-IQ4\_XS) $\approx$340 GiB, all within a $\pm7\%$ band around $\approx$320
GiB.  The ranking of detection scores is therefore not explained by quantized
memory footprint: GLM-5.2 is the heaviest yet ranks second, while MiniMax-M3 is
the lightest yet ranks third.  This indicates that the primary Kimi backbone
used in the main paper is not benefiting from a footprint advantage but from its
representation quality under the LAPE protocol.

\paragraph{Within-backbone: Enriched over Plain.}
Under both protocols, Enriched is at least as strong as Plain for every
backbone except MiniMax on ADReSSo CV (89.15 vs.\ 89.78 F1), where the gap is
only $-0.63$.  The Enriched-over-Plain F1 gains are largest for Kimi (ADReSS
CV $+4.55$, LOSO $+5.56$; ADReSSo CV $+1.82$, LOSO $+3.61$) and for GLM on
ADReSS (CV $+2.01$, LOSO $+4.62$).  MiniMax shows the smallest and least
consistent gains (ADReSS CV $+1.90$ / LOSO $+0.95$; ADReSSo CV $-0.63$ / LOSO
$+1.78$).  The two design choices thus transfer to all three backbones.

\paragraph{Across backbones: scale and performance.}
Absolute scores rank Kimi $>$ GLM $>$ MiniMax on both benchmarks and protocols.
This ordering is monotone in total parameter count (1T $>$ 744B $>$ 428B) but
not in per-token activated parameters, since GLM-5.2 activates 40B per token
versus Kimi's 32B yet trails Kimi by 1.82--4.81 F1 across the four settings.
Per-token activation therefore does not predict AD-detection ranking here; the
gain from 23B (MiniMax) to 32B (Kimi) activated capacity is larger than the
further increase to 40B (GLM).  One interpretation is that total MoE capacity
governs the richness of the instruction-conditioned text representation, while
per-token activation governs inference cost rather than representation quality
on this small-token, single-utterance task.  Combined with the quantized
footprint result above, the Enriched-over-Plain direction is robust across the
three parameter scales and quantization levels tested here.